\pdfoutput=1

\documentclass{article}

\usepackage[final]{corl_2019} 

\usepackage{balance}
\usepackage[utf8]{inputenc} 
\usepackage[T1]{fontenc}    
\usepackage{hyperref}       
\usepackage{url}            
\usepackage{booktabs}       
\usepackage{amsfonts}       
\usepackage{nicefrac}       
\usepackage{microtype}      

\usepackage{amsmath,amsfonts,bm}

\def\eqref#1{equation~\ref{#1}}

\def\1{\bm{1}}

\def\rtau{{\bm{\tau}}}

\DeclareMathAlphabet{\mathsfit}{\encodingdefault}{\sfdefault}{m}{sl}
\SetMathAlphabet{\mathsfit}{bold}{\encodingdefault}{\sfdefault}{bx}{n}

\newcommand{\E}{\mathbb{E}}

\usepackage{times}
\usepackage{graphicx} 
\usepackage{natbib}

\usepackage[font=footnotesize,labelfont=footnotesize]{subcaption}
\usepackage[font=footnotesize,labelfont=footnotesize]{caption}

\usepackage{wrapfig}
\usepackage{url}
\usepackage{mathtools}
\usepackage{float}

\usepackage{lipsum}
\usepackage{xcolor}

\title{MAME : Model-Agnostic Meta-Exploration}

\author{
  Swaminathan Gurumurthy\thanks{Robotics Institute, Carnegie Mellon University, Pittsburgh, PA 15213, USA. Corresponding authors: sgurumur@cs.cmu.edu, skumar2@cs.cmu.edu }\\
  \And
  Sumit Kumar\footnotemark[1] \\
  \And
  Katia Sycara\footnotemark[1] \\
}

\begin{document}
\maketitle


\begin{abstract}
Meta-Reinforcement learning approaches aim to develop learning procedures that can adapt quickly to a distribution of tasks with the help of a few examples. Developing efficient exploration strategies capable of finding the most useful samples becomes critical in such settings. Existing approaches towards finding efficient exploration strategies add auxiliary objectives to promote exploration by the pre-update policy, however, this makes the adaptation using a few gradient steps difficult as the pre-update (exploration) and post-update (exploitation) policies are often quite different. Instead, we propose to explicitly model a separate exploration policy for the task distribution. Having two different policies gives more flexibility in training the exploration policy and also makes adaptation to any specific task easier. We show that using self-supervised or supervised learning objectives for adaptation allows for more efficient inner-loop updates and also demonstrate the superior performance of our model compared to prior works in this domain. 
\end{abstract}



\section{Introduction}
\vspace{-5pt}
Reinforcement learning (RL) approaches have seen many successes in recent years, from mastering the complex game of Go~\citep{silver2017mastering} to even discovering molecules~\citep{olivecrona2017molecular}. However, a common limitation of these methods is their propensity to overfitting on a single task and inability to adapt to even slightly perturbed configuration~\citep{zhang2018study}. On the other hand, humans have this astonishing ability to learn new tasks in a matter of minutes by using their prior knowledge and understanding of the underlying task mechanics. Drawing inspiration from human behaviors, researchers have proposed to incorporate multiple inductive biases and heuristics to help the models learn quickly and generalize to unseen scenarios. However, despite a lot of effort it has been difficult to approach human levels of data efficiency and generalization. 

Meta reinforcement learning addresses these shortcomings by learning how to learn these inductive biases and heuristics from the data itself. It strives to learn an algorithm that allows an agent to succeed in a previously unseen task or environment when only limited experience is available. These inductive biases or heuristics can be induced in the model in various ways like optimization algorithm, policy, hyperparameters, network architecture, loss function, exploration strategies \cite{sharaf2019meta}, \textit{etc}. Recently, a class of parameter initialization based meta-learning approaches have gained attention like Model Agnostic Meta-Learning (MAML)~\citep{finn2017model}. MAML finds a good initialization for a model or a policy that can be adapted to a new task by fine-tuning with policy gradient updates from a few samples of that task.

Since the objective of meta-RL algorithms is to adapt to a new task from a few examples, efficient exploration strategies are crucial for quickly finding the optimal policy in a new environment. Some recent works have tried to address this problem by improving the credit assignment of the meta learning objective to the pre-update trajectory distribution \citep{stadie2018some, rothfuss2018promp}. However, that requires transitioning the base policy from exploration behavior to exploitation behavior using one or few policy gradient updates. This limits the applicability of these methods to cases where the post-update (exploitation) policy is similar to the pre-update (exploration) policy and can be obtained with only a few updates. Additionally, for cases where pre- and post-update policies are expected to exhibit different behaviors, large gradient updates may result in training instabilities and poor performance at convergence. 

To address this problem, we propose to explicitly model a separate exploration policy for the task distribution. The exploration policy is trained to find trajectories that can lead to fast adaptation of the exploitation policy on the given task. This formulation provides much more flexibility in training the exploration policy. We'll also show that separating the exploration and exploitation policy helps us improve/match the performance of the baseline approaches on meta-RL benchmark tasks.
We further show that, in order to perform stable and fast adaptation to the new task, it is often more useful to use self-supervised or supervised learning approaches to perform the inner loop/meta updates, where possible. This also helps obtain more stable gradient update steps while training exploitation policy with the trajectories collected from a different exploration policy.
\vspace{-10pt}

\section{Related work}
\vspace{-5pt}

Meta-learning algorithms proposed in the RL community include approaches based on recurrent models \citep{duan2016rl,finn2017meta}, metric learning \citep{snell2017prototypical,sung2018learning}, and learning optimizers \citep{alex20181storder}. Recently, \citet{finn2017model} proposed Model Agnostic Meta-Learning (MAML) which aims to learn a policy that can generalize to a distribution of tasks. Specifically, it aims to find a good initialization for a policy that can be adapted to any task sampled from the distribution by fine-tuning with policy gradient updates from a few samples of that task. 

Efficient exploration strategies are crucial for finding trajectories that can lead to quick adaptation of the policy in a new environment. Recent works \cite{gupta2018meta, rakelly2019efficient} have proposed structured exploration strategies using latent variables to perform efficient exploration across successive episodes, however, they did not explicitly incentivize exploration in pre-update episodes.
E-MAML \citep{stadie2018some} made the first attempt at assigning credit for the final expected returns to the pre-update distribution in order to incentivize exploration in each of the pre-update episodes.  \citet{rothfuss2018promp} proposed Proximal Meta-Policy search (ProMP) where they incorporated the causal structure for more efficient credit assignment and proposed a low variance curvature surrogate objective to control the variance of the corresponding policy gradient update. However, these methods make use of a single base policy for both exploration and exploitation while relying on one or few gradient updates to transition from the exploration behavior to exploitation behavior. Over the next few sections, we illustrate that this approach is problematic and insufficient when the exploration and exploitation behaviors are quite different from each other.

A number of prior works have tried to utilize self-supervised objectives \cite{mahmoudieh2017self, florensa2019self, kahn2018self, pathak2017curiosity, wortsman2018learning} to ease learning especially when the reward signal itself is insufficient to provide the required level of feedback. Drawing inspiration from these approaches, we modify the inner loop update/adaptation step in MAML using a self-supervised objective to allow more stable and faster updates. Concurrent to our work, \citet{yang2019norml} also decoupled exploration and adaptation policies where the latter is initialized as a learnable offset to the exploration policy.  

\vspace{-10pt}
\section{Background}

\vspace{-5pt}
\subsection{Meta-Reinforcement Learning}

Unlike RL which tries to find an optimal policy for a single task, meta-RL aims to find a policy that can generalize to a distribution of tasks. Each task $\mathcal{T}$ sampled from the distribution $\rho(\mathcal{T})$ corresponds to a different Markov Decision Process (MDP) defined by the tuple $(\mathcal{S},\mathcal{A},\boldsymbol{P},r,\gamma,H)$ with state space $\mathcal{S}$, action space $\mathcal{A}$, transition dynamics $\boldsymbol{P}$, reward function $r$, discount factor $\gamma$, and time horizon $H$. These MDPs are assumed to have similar state and action space but might differ in the reward function $r$ or the environment dynamics $\boldsymbol{P}$. The goal of meta RL is to quickly adapt the policy to any task $\mathcal{T} \sim \rho(\mathcal{T})$ with the help of few examples from that task. 
\vspace{-5pt}
\subsection{Credit Assignment in Meta-RL}

MAML is a gradient-based meta-RL algorithm that tries to find a good initialization for a policy which can be adapted to any task $\mathcal{T} \sim \rho(\mathcal{T})$ by fine-tuning with one or more gradient updates using the sampled trajectories of that task. MAML maximizes the following objective function:
\begin{align}
J(\theta) = \E_{\mathcal{T} \sim \rho({\mathcal{T}})} \big[ \E_{\rtau' \sim P_{\mathcal{T}}(\rtau' | \theta')}\left[ R(\rtau') \right]  \big] \quad \nonumber\\
\text{with} \quad \theta':=U(\theta, \mathcal{T})= \theta + \alpha \nabla_\theta \E_{\rtau \sim P_{\mathcal{T}}(\rtau|\theta)} \left[R(\rtau)\right]
\label{meta-loss-maml}
\end{align}
where $U$ is the update function that performs one policy gradient ascent step to maximize the expected reward $R(\rtau)$ obtained on the trajectories $\rtau$ sampled from task $\mathcal{T}$.

\citet{rothfuss2018promp} showed that the gradient of the objective function $J(\theta)$ in Eq. \ref{meta-loss-maml} can be written as:
\begin{align*}
\nabla_\theta J(\theta) = \E_{\mathcal{T} \sim \rho(\mathcal{T})} \Bigg[ \E_{\substack{\rtau \sim P_{\mathcal{T}}(\rtau | \theta) 
\rtau' \sim P_{\mathcal{T}}(\rtau' | \theta')}} \bigg[ \nabla_\theta J_{\text{post}}(\rtau, \rtau') 
+ \nabla_\theta J_{\text{pre}}\left(\rtau, \rtau'\right) \bigg] \Bigg]
\end{align*}
where, 
\begin{align}\label{eq:j_post}
\nabla_\theta J_{\text{post}}(\rtau, \rtau') &= \underbrace{\nabla_{\theta'} \log \pi_\theta(\rtau') R(\rtau')}_{\nabla_{\theta'} J^{\text{outer}}} \underbrace{\left(I + \alpha  R(\rtau) \nabla_\theta^2 \log \pi_{\theta}(\rtau) \right)}_{\text{transformation from $\theta'$ to $\theta$}}\\ 
\nabla_\theta J_{\text{pre}}(\rtau, \rtau') &= \alpha \nabla_\theta \log \pi_\theta(\rtau) \bigg( \underbrace{  (\nabla_\theta \log \pi_\theta(\rtau) R(\rtau))^\top }_{\nabla_\theta J^{\text{inner}}}\underbrace{ (\nabla_{\theta'} \log \pi_{\theta'}(\rtau') R(\rtau') ) }_{\nabla_{\theta'} J^{\text{outer}}} \bigg)\label{eq:j_pre}
\end{align}

The first term $\nabla_\theta J_{\text{post}}(\rtau, \rtau')$ corresponds to a policy gradient step on the post-update policy $\pi_{\theta'}$ w.r.t.. the post-update parameters $\theta'$ which is then followed by a linear transformation from $\theta'$ to $\theta$ (pre-update parameters). Note that $\nabla_\theta J_{\text{post}}(\rtau, \rtau')$ optimizes $\theta$ to increase the likelihood of the trajectories $\rtau^\prime$ that lead to higher returns given some trajectories $\rtau$. However, this term does not optimize $\theta$ to yield trajectories $\rtau$ that lead to good adaptation steps. That is, infact, done by the second term $\nabla_\theta J_{\text{pre}}(\rtau, \rtau')$. It optimizes for the pre-update trajectory distribution, $P_{\mathcal{T}}(\rtau | \theta)$, i.e, increases the likelihood of trajectories $\rtau$ that lead to good adaptation steps.

During optimization, MAML only considers $J_{\text{post}}(\rtau, \rtau')$ and ignores $J_{\text{pre}}(\rtau, \rtau')$. Thus MAML finds a policy that adapts quickly to a task given relevant experiences, however, the policy is not optimized to gather useful experiences from the environment that can lead to fast adaptation. 

\citet{rothfuss2018promp} proposed ProMP where they analyze this issue with MAML and incorporate $\nabla_\theta J_{\text{pre}}(\rtau, \rtau')$ term in the update as well. They used The Infinitely Differentiable Monte-Carlo Estimator (DICE)~\citep{foerster2018DICE} to allow causal credit assignment on the pre-update trajectory distribution, however, the gradients computed by DICE still have high variance. To remedy this, they proposed a low variance (and slightly biased) approximation of the DICE based loss that leads to stable updates. 

\vspace{-10pt}
\section{Method}
\vspace{-5pt}

The pre-update and post-update policies are often expected to exhibit very different behaviors, i.e, exploration and exploitation behaviors respectively. For instance, consider a 2D environment where a task corresponds to reaching a goal location sampled randomly from a semi-circular region (example shown in appendix). The agent receives a reward only if it lies in some vicinity of the goal location. The optimal pre-update or exploration policy is to move around in the semi-circular region whereas the ideal post-update or exploitation policy will be to reach the goal state as fast as possible once the goal region is discovered. Clearly, the two policies are expected to behave very differently. In such cases, transitioning a single policy from pure exploration phase to pure exploitation phase via policy gradient updates will require multiple steps. Unfortunately, this significantly increases the computational and memory complexities of the algorithm. Furthermore, it may not even be possible to achieve this transition via few gradient updates. This raises an important question: \textsc{Do we really need to use the pre-update policy for exploration as well? Can we use a separate policy for exploration?}

\textbf{Using separate policies for pre-update and post-update sampling:}
The straightforward solution to the above problem is to use a separate exploration policy $\mu_{\phi}$ responsible for collecting trajectories for the inner loop updates to get $\theta'$. Following that, the post-update policy $\pi_{\theta'}$ can be used to collect trajectories for performing the outer loop updates. Unfortunately, this is not as simple as it sounds. To understand this, let's look at the inner loop updates:
\begin{align*}
U(\theta, \mathcal{T}) = \theta + \alpha \nabla_\theta \E_{\rtau \sim P_{\mathcal{T}}(\rtau|\theta)} \left[R(\rtau)\right]
\end{align*}

When the exploration policy is used for sampling trajectories, we need to perform importance sampling. The update would thus become:
\begin{align*}
    U(\theta, \mathcal{T}) = \theta + \alpha \nabla_\theta \E_{\rtau \sim Q_{\mathcal{T}}(\rtau|\phi)} \left[ \frac{P_{\mathcal{T}}(\rtau|\theta)}{Q_{\mathcal{T}}(\rtau|\phi)}  R(\rtau)\right]
\end{align*}
where $P_{\mathcal{T}}(\rtau|\theta)$ and $Q_{\mathcal{T}}(\rtau|\phi)$ represent the trajectory distribution sampled by $\pi_{\theta}$ and $\mu_{\phi}$ respectively. Note that the above update is an off-policy update which results in high variance estimates when the two trajectory distributions are quite different from each other. This makes it infeasible to use the importance sampling update in the current form. 
In fact, this is a more general problem that arises even in the on-policy regime. The policy gradient updates in the inner loop results in both $\nabla_\theta J_{\text{pre}}$ and $\nabla_\theta J_{\text{post}}$ terms being high variance. This stems from the mis-alignment of the outer gradients ($\nabla_{\theta'} J^{outer}$) and the inner gradient, hessian $\left(\nabla_{\theta} J^{\text{inner}}, \nabla_\theta^2 \log \pi_{\theta}(\rtau)\right)$ terms appearing in Eq. \ref{eq:j_post} and \ref{eq:j_pre}.
This motivates our second question: \textsc{Do we really need the pre-update gradients to be policy gradients? Can we use a different objective in the inner loop to get more stable updates?}

\textbf{Using a self-supervised/supervised objective for the inner loop update step:}
The instability in the inner loop updates arises due to the high variance nature of the policy gradient update. Note that the objective of inner loop update is to provide some task specific information to the agent with the help of which it can adapt its behavior in the new environment. 
We believe that this could be achieved using some form of self-supervised or supervised learning objective in place of policy gradient in the inner loop to ensure that the updates are more stable. We propose to use a network for predicting some task (or MDP) specific property like reward function, expected return or value. During the inner loop update, the network updates its parameters by minimizing its prediction error on the given task. Unlike prior meta-RL works where the task adaptation in the inner loop is done by policy gradient updates, here, we update some parameters shared with the exploitation policy using a supervised loss objective function which leads to stable updates during the adaptation phase. However, note that the variance and usefulness of the update depends heavily on the choice of the self-supervision/supervision objective. We delve into this in more detail in Section \ref{sec:self-sup}.

\vspace{-5pt}
\subsection{Model}
\vspace{-5pt}
\begin{wrapfigure}{r}{0.6\textwidth}
    \centering
    \includegraphics[width=0.6\columnwidth]{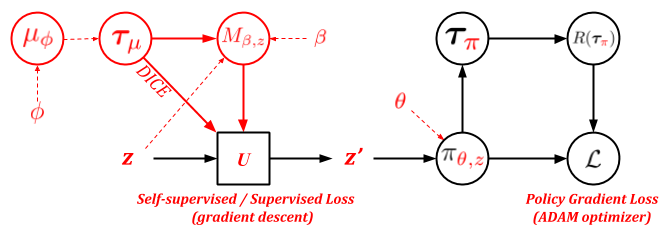}
    \caption{Model Flowchart: Black structures are those consistent with E-MAML/ProMP. Red structures are the key differences with E-MAML/ProMP (The thin-dotted arrow means the parameters related to that node.). Specifically, the pre-update trajectories $\rtau$ are now collected using a separate exploration policy $\mu_{\phi}$. The corresponding adaptation update is performed using a self-supervised/supervised objective, $\left(M_{\beta,z}(s,a)-\overline{M}(s,a)\right)^2$, on $z$ to give $z'$ and the policy $\pi_{\theta, z'}$ is parameterized using the task specific parameters $z'$ and the task agnostic parameters $\theta$ }
    \label{fig:flowchart}
\end{wrapfigure}

Our proposed model comprises of three modules, the exploration policy $\mu_{\phi}(s)$, the exploitation policy $\pi_{\theta,z}(s)$, and the self-supervision network $M_{\beta,z}(s,a)$. Note that $M_{\beta,z}$ and $\pi_{\theta,z}$ share a set of parameters $z$ while containing their own set of parameters $\beta$ and $\theta$ respectively. We describe our proposed model in Fig. \ref{fig:flowchart}. 
Our model differs from E-MAML/ProMP because of the separate exploration policy, the separation of task-specific parameters $z$ and task agnostic parameters $\theta$, and the self-supervised update as shown in Fig. \ref{fig:flowchart}.

The agent first collects a set of trajectories $\rtau$ using its exploration policy $\mu_{\phi}$ for each task $\mathcal{T} \sim \rho(\mathcal{T})$. It then updates the shared parameter $z$ by minimizing the regression loss $\left(M_{\beta,z}(s,a)-\overline{M}(s,a)\right)^2$ on the sampled trajectories $\rtau$:
\begin{align}\label{eq:update_z}
    z'=  U(z, \mathcal{T}) = z - \alpha \nabla_z \E_{\rtau \sim 
    Q_{\mathcal{T}}(\rtau|\phi)}\left[\sum_{t=0}^{H-1}\left(M_{\beta,z}(s_t,a_t)-\overline{M}(s_t,a_t)\right)^2 \right]
\end{align}
where, $\overline{M}(s,a)$ is the target, which can be any task specific quantity like reward, return, value, next state etc. After obtaining the updated parameters $z'$ for each task $\mathcal{T}$, the agent samples the (validation) trajectories $\rtau'$ using its updated exploitation policy $\pi_{\theta,z'}$. Effectively, $z'$ encodes the necessary information regarding the task that helps an agent in adapting its behavior to maximize its expected return whereas $\theta$ remain task agnostic. A similar approach was proposed by \citet{zintgraf2018caml} to learn task-specific behavior using context variable with MAML. 

The collected trajectories are then used to perform a policy gradient update to all parameters $z, \theta, \phi$ and $\beta$ using the following objective:
\begin{align}\label{eq:outer_loop}
J(z', \theta) = \E_{\mathcal{T} \sim \rho({\mathcal{T}})} \big[ \E_{\rtau_{\pi}^{\mathcal{T}} \sim P_{\mathcal{T}}(\rtau_{\pi}^{\mathcal{T}} | \theta, z')}\left[ R(\rtau_{\pi}^{\mathcal{T}}) \right]  \big]
\end{align}

In order to allow multiple outer-loop updates, we use the PPO \cite{schulman2017proximal} objective instead of the vanilla policy gradient objective to maximize Eq. \ref{eq:outer_loop}. Furthermore, we don't perform any outer loop updates on $z$ and treat it as a shared latent variable with fixed initial values of $0$ as proposed in \cite{zintgraf2018caml}. The reason being, that the bias term in the layers connecting $z$ to the respective networks would learn to compensate for the initialization. We only update $z$ to $z'$ in the inner loop to obtain a task specific latent variable. 

Note that in all the prior meta reinforcement learning algorithms, both the inner-loop update and the outer-loop update are policy gradient updates. In contrast, in this work, the inner-loop update is a supervised learning gradient descent update whereas the outer-loop update remains a policy gradient update. 

The outer loop gradients w.r.t. $\phi$, $\nabla_{\phi} J(z',\theta)$ can be simplified by multiplying the DICE \cite{foerster2018DICE} operator inside the expectation in Eq. \ref{eq:update_z} as proposed by \citet{rothfuss2018promp}. This allows the gradients w.r.t. $\phi$ to be computed in a straightforward manner with back-propagation. This also eliminates the need to apply the policy gradient trick to expand Eq. \ref{eq:update_z} for gradient computation. The inner loop update then becomes:

\begin{align*}
    z' =  U(z, \mathcal{T}) = z - \alpha \nabla_z \E_{\rtau \sim 
    Q_{\mathcal{T}}(\rtau|\phi)}\left[\sum_{t=0}^{H-1}\left( \prod_{t'=0}^t \frac{\mu_{\phi}(a_{t'}|s_{t'})}{\bot(\mu_{\phi}(a_{t'}|s_{t'}))} \right)\left(M_{\beta,z}(s_t,a_t)-\overline{M}(s_t,a_t)\right)^2 \right]
\end{align*}
where $\bot$ is the stop gradient operator as introduced in \cite{foerster2018DICE}.

The pseudo-code of our algorithm is shown in appendix (see algorithm \ref{alg:multibot}). 
However, we found that implementing algorithm \ref{alg:multibot} as it is, using DICE leads to high variance gradients for $\phi$, resulting in instability during training and poor performance of the learned model. To understand this, let's look at the vanilla DICE gradients for the exploration parameters $\phi$, which can be written as follows:
\begin{align*}
    \nabla_\phi J(z', \theta) = \E_{\mathcal{T} \sim \rho(\mathcal{T})} \Bigg[ \E_{\rtau \sim 
    Q_{\mathcal{T}}(\rtau|\phi)} \sum_{t=0}^{H-1} \alpha \nabla_\phi \log \mu_\phi(s_t) 
    \bigg[\sum_{t'=t}^{H-1} &\Big( \E_{\rtau' \sim 
    P_{\mathcal{T}}(\rtau'|\theta,z')} \left(\nabla_{z'} \log \pi_{\theta, z'}(\rtau') R(\rtau')\right)^\top \Big) \\
   & \left(\nabla_{z} \left(M_{\beta,z}(s_t,a_t)-\overline{M}(s_t,a_t)\right)^2 \right) \bigg] \Bigg]
\end{align*}

The above expression can be viewed as a policy gradient update: 
\begin{align}\label{eq:gradj_ru}
    \nabla_\phi J(z', \theta) =\E_{\mathcal{T} \sim \rho(\mathcal{T})} \bigg[ \E_{\rtau \sim Q_{\mathcal{T}}(\rtau|\phi)}\sum_{t=0}^{H-1} \alpha \nabla_\phi \log \mu_\phi(s_t) R_t^{\mu} \bigg]
\end{align}

with returns  
\begin{align}\label{eq:returns}
R_t^{\mu}= \Bigg[\sum_{t'=t}^{H-1} & \Big( \E_{\rtau' \sim 
        P_{\mathcal{T}}(\rtau'|\theta,z')} \left(\nabla_{z'} \log \pi_{\theta, z'}(\rtau') R(\rtau')\right)^\top \Big) \left(\nabla_{z} \left(M_{\beta,z}(s_t,a_t)-\overline{M}(s_t,a_t)\right)^2 \right) \Bigg]
\end{align}

Note that the variance depends on the policy gradient terms computed in the outer-loop and the choice of self-supervision. We'll explain the latter in Sec \ref{sec:self-sup}. However, irrespective of the choice, we can use value function based variance reduction (\cite{mnih2016asynchronous}) by substituting the above computed returns with advantages, i.e, we replace the return $R_t^{\mu}$ in Eq. \ref{eq:returns} with an advantage estimate $A_t^{\mu}$ and use a PPO (\cite{schulman2017proximal}) objective to allow multiple outer loop updates: :

\begin{equation*}
    \nabla_\phi \hat{J}(z', \theta) =\E_{\mathcal{T} \sim \rho(\mathcal{T})} \bigg[ \E_{\rtau \sim 
        Q_{\mathcal{T}}(\rtau|\phi_o)}  \left[\sum_{t=0}^{H-1} \alpha \nabla_{\phi} \min \left(\frac{\mu_\phi(s_t)}{\mu_{\phi_{o}}(s_t)} A^{\mu}_t ~, ~ \text{clip}^{1+\epsilon}_{1-\epsilon} \left(\frac{\mu_\phi(s_t)}{\mu_{\phi_{o}}(s_t)}\right) A^{\mu}_t   \right)\right]
\end{equation*}
where, 
\begin{align} \label{eq:advantage}
    A_t^{\mu} = r_t^{\mu} + V_{t+1}^{\mu} - V_{t}^{\mu} 
\end{align}
where $V_{t}^{\mu}$ is computed using a neural network or a linear feature baseline \citep{duan2016benchmarking} fitted on the returns $R_t^{\mu}$. where $r_t^{\mu}$ is given by:
\begin{align} \label{eq:rewards_exp}
r_t^{\mu} = &\Big( \E_{\rtau' \sim 
        P_{\mathcal{T}}(\rtau'|\theta,z')} \left(\nabla_{z'} \log \pi_{\theta, z'}(\rtau') R(\rtau')\right)^\top \Big) \left(\nabla_{z} \left(M_{\beta,z}(s_t,a_t)-\overline{M}(s_t,a_t)\right)^2 \right)
\end{align}

\subsubsection{Self-Supervised/Supervised Objective} \label{sec:self-sup}
It is important to note that the self-supervised/supervised learning objective not only guides the adaptation step but also influences the exploration policy update as seen in Eq. \ref{eq:gradj_ru} and \ref{eq:returns}. We mentioned above that the self-supervised/supervised learning objective could be as simple as a value/reward/return/next state prediction for each state (state-action pair). 
However, the exact choice of the objective can be critical to the final performance and stability. From the perspective of the adaptation step, the only criterion is that the self-supervised objective should contain enough task specific information to allow a useful adaptation step. 
For example, it would not be a good idea to use the rewards self-supervision in sparse/noisy reward scenarios or the next state predictions as self-supervision when the dynamics model does not change much over tasks since the self-supervision updates in such cases will not carry enough task specific information. From the perspective of the exploration policy updates, an additional requirement would be to ensure that the returns computed in Eq. \ref{eq:returns} are low variance and unbiased, which further translates to saying $\nabla_{z} \left(M_{\beta,z}(s_t,a_t)-\overline{M}(s_t,a_t)\right)^2$ should ideally be low variance and unbiased. For example, using the cumulative future returns as self-supervision might lead to a very high variance update in certain environments. Thus, finding a generalizable self-supervision/supervision objective which satisfies both properties mentioned above across all scenarios is a challenging task. 

In our experiments, we found that, using $N$-step return prediction for supervision works reasonably well across all the environments. 
This acts as a trade-off between predicting the full return (which was high variance but also more task-specific info) and the reward (which was lower variance but lower task-specific info). Hence, $\overline{M}(s_t,a_t)$ becomes $\overline{M}(s_t,a_t, s_{t+1}, a_{t+1},..... s_{t+N-1}, a_{t+N-1}) = \sum_{t'=t}^{t+N-1} r(s_t',a_t') $. However, using $M_{\beta,z}$ to directly predict $\overline{M}$ would still lead to high variance in $\nabla_{z} \left(M_{\beta,z}(s_t,a_t)-\overline{M}\right)^2$.
Thus, we use the truncated $N$-step successor representations \citep{kulkarni2016deep} (similar to N-step returns) $m_{\beta} (s_t, a_t)$ and a linear layer on top of that to compute $M_{\beta,z}(s_t, a_t) = w_{\beta}^T m_{\beta}(s_t, a_t)$. Using the successor representations can effectively be seen as using a more accurate/powerful baseline than directly predicting the N-step returns using the $(s_t, a_t)$ pair. 

\vspace{-10pt}

\section{Experiments}
\vspace{-5pt}
We evaluate our proposed model on a set of $6$ benchmark continuous control environments, \texttt{\texttt{Ant-Fwd-Bwd}}, \texttt{Half-Cheetah-Fwd-Bwd}, \texttt{Half-Cheetah-Vel}, \texttt{Walker2D-Fwd-Bwd}, \texttt{Walker2D-Rand-Params} and \texttt{Hopper-Rand-Params} used in \cite{rothfuss2018promp}. We also compare our method with $3$ baseline approaches: MAML, EMAML and ProMP. Furthermore, we also perform ablation experiments to analyze different components and design choices of our model on a toy $2$D point environment proposed by \cite{rothfuss2018promp}.

The details of the network architecture and the hyperparameters used for learning have been mentioned in the appendix. We would like to state that we have not performed much hyperparameter tuning due to computational constraints and we expect the results of our method to show further improvements with further tuning. Also, we restrict ourselves to a single adaptation step in all environments for the baselines as well as our method, but it can be easily extended to multiple gradient steps as well by conditioning the exploration policy on the latent parameters $z$. 

The results of the baselines for the benchmark environments have been borrowed directly from the the official ProMP website \footnote{\url{https://sites.google.com/view/pro-mp/experiments}}. For the point environments, we have used their official implementation\footnote{\url{https://github.com/jonasrothfuss/ProMP}}.

\begin{figure}[h]
    \begin{minipage}{.33\textwidth}
        \centering
        \includegraphics[width=\textwidth]{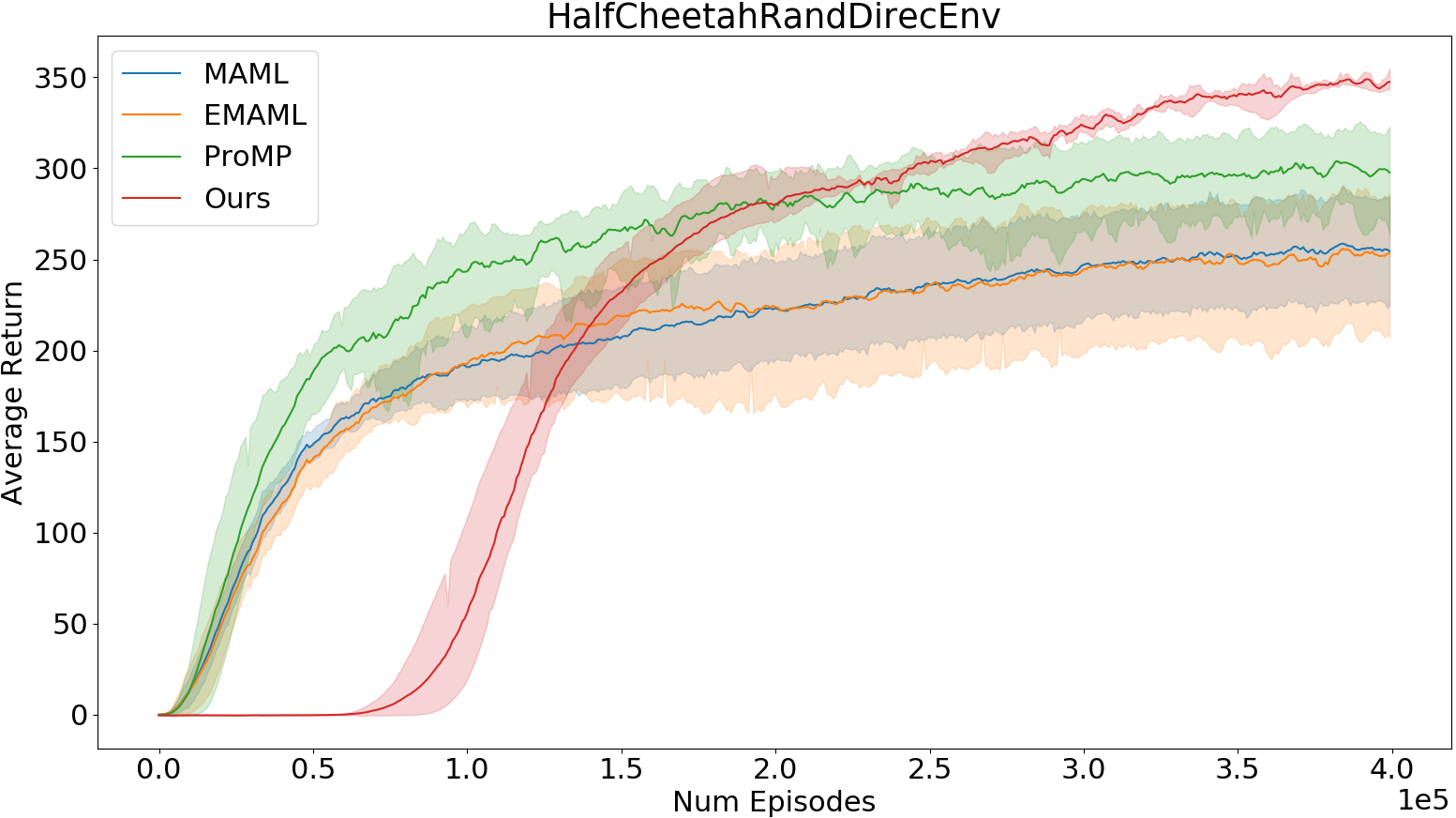} 
        \subcaption{\texttt{Half-Cheetah-Fwd-Bwd}}
        \label{fig:pointenv}
    \end{minipage}\hfill
    \begin{minipage}{.33\textwidth}
        \centering
        \includegraphics[width=\textwidth]{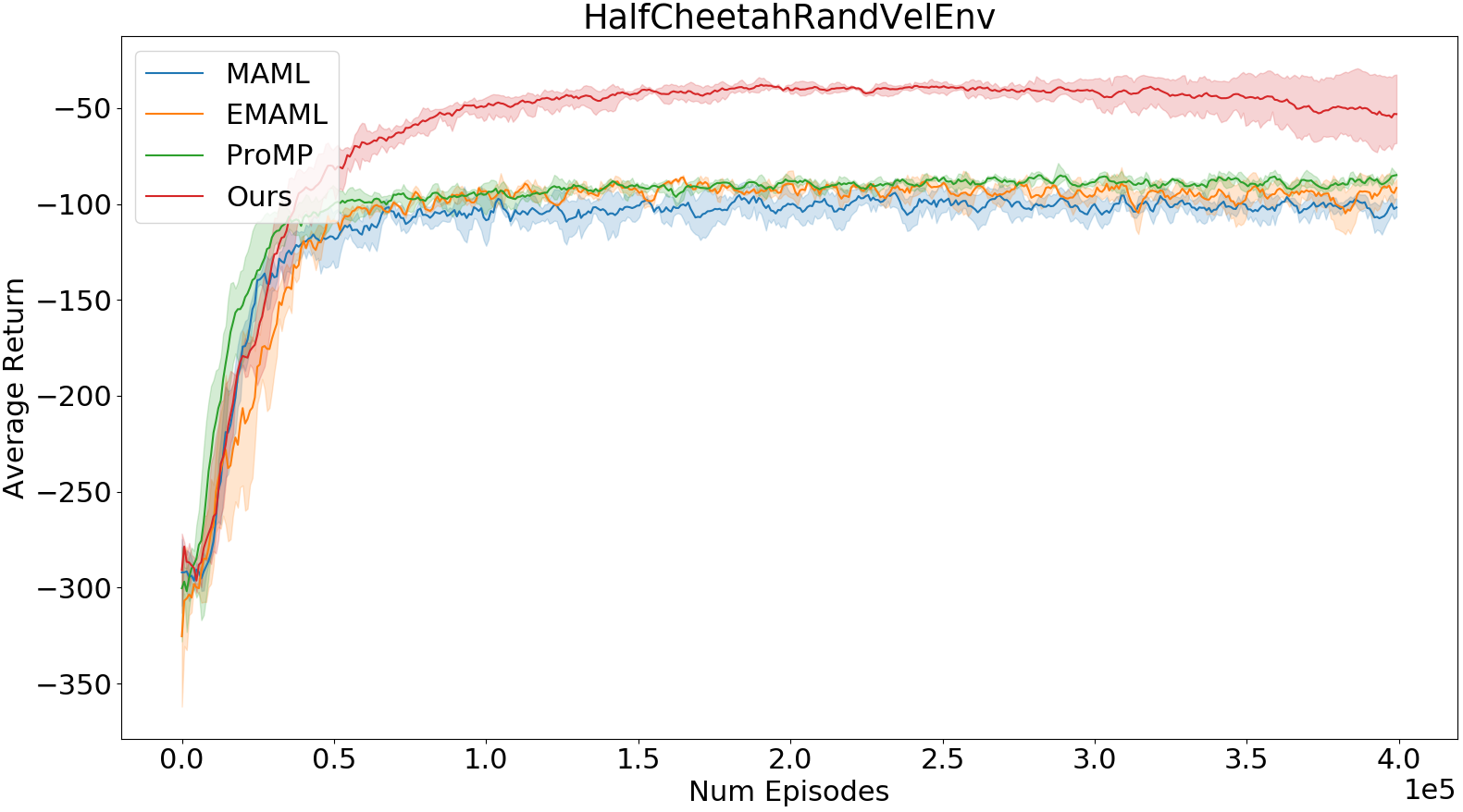} 
        \subcaption{\texttt{Half-Cheetah-Vel}}
        \label{fig:antdir}
    \end{minipage}
    \begin{minipage}{.33\textwidth}
        \centering
        \includegraphics[width=\textwidth]{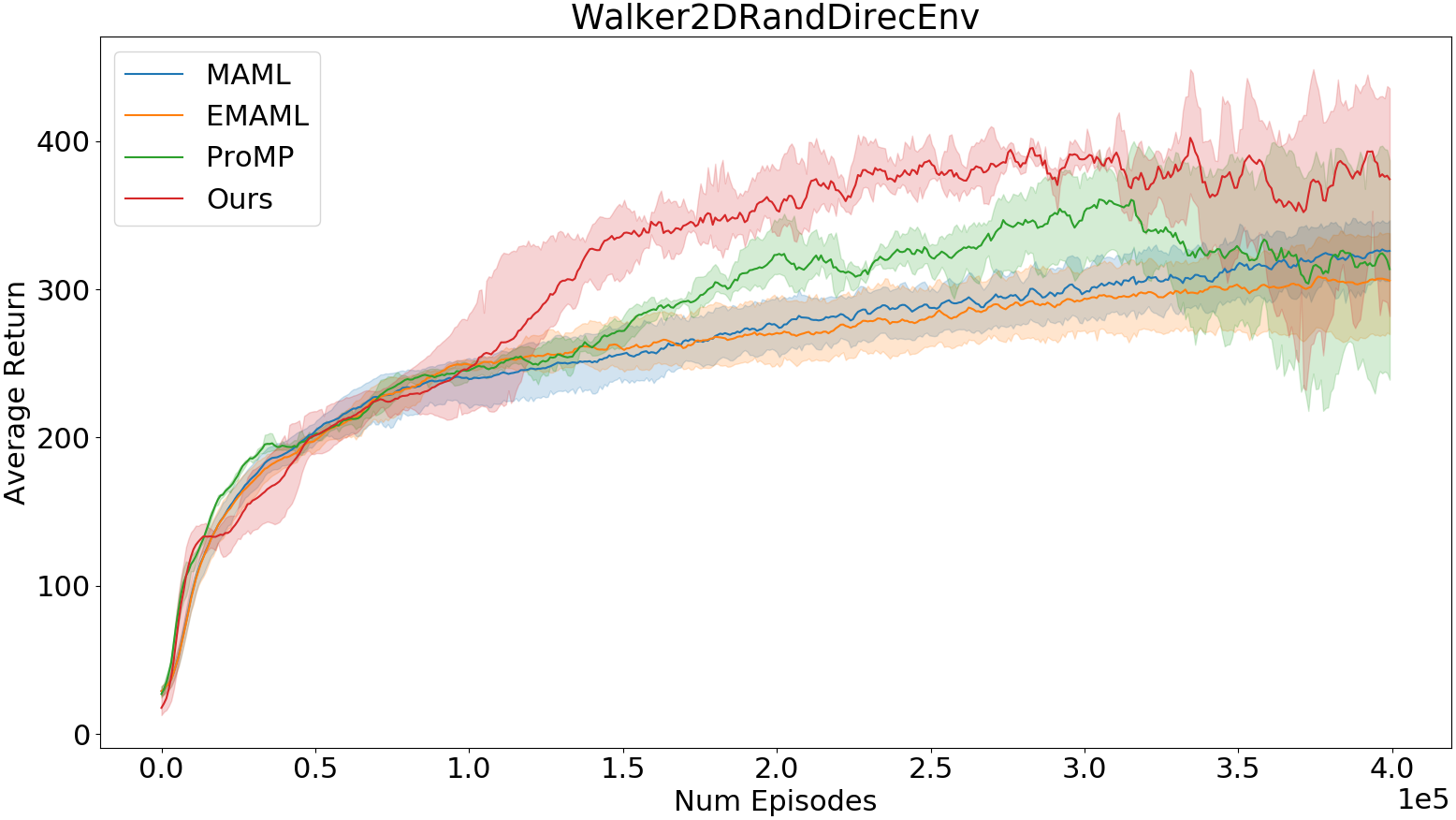} 
        \subcaption{\texttt{Walker2D-Fwd-Bwd}}
        \label{fig:hcdir}
    \end{minipage}

    \begin{minipage}{.33\textwidth}
        \centering
        \includegraphics[width=\textwidth]{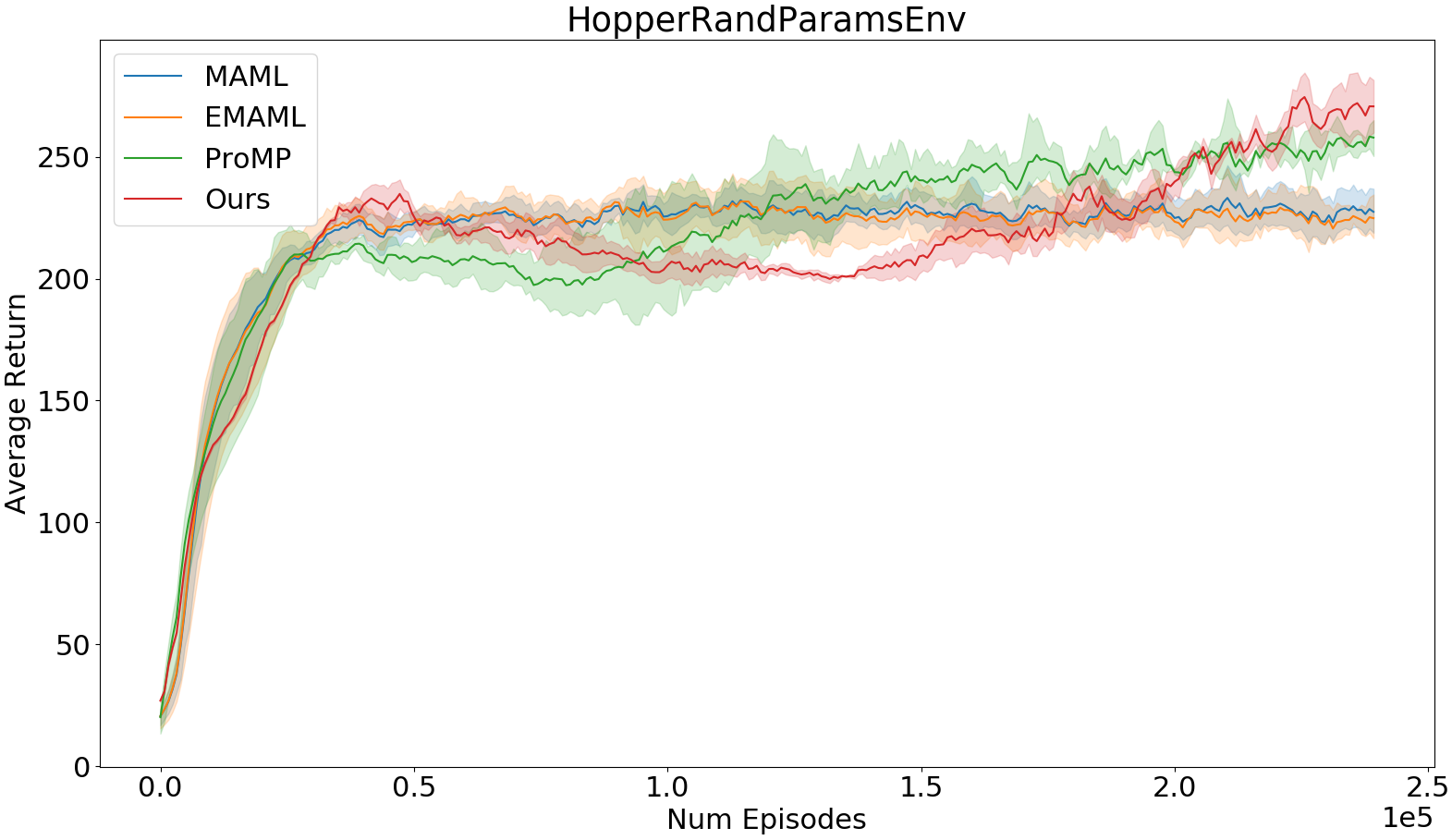} 
        \subcaption{\texttt{Hopper-Rand-Params}}
        \label{fig:hcvel}
    \end{minipage}
    \begin{minipage}{.33\textwidth}
        \centering
        \includegraphics[width=\textwidth]{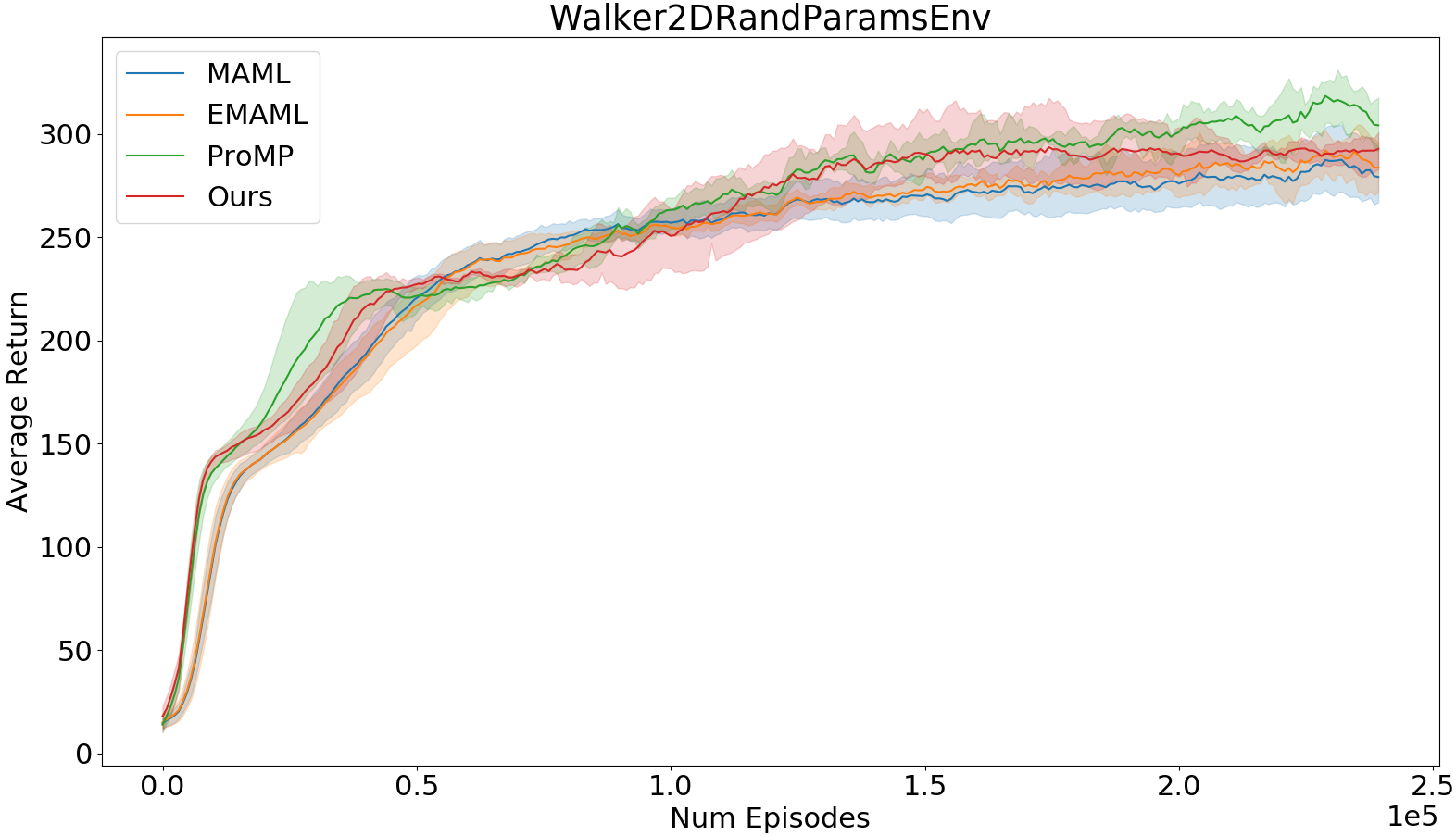} 
        \subcaption{\texttt{Walker2D-Rand-Params}}
        \label{fig:hcvel}
    \end{minipage}
    \begin{minipage}{.33\textwidth}
        \centering
        \includegraphics[width=\textwidth]{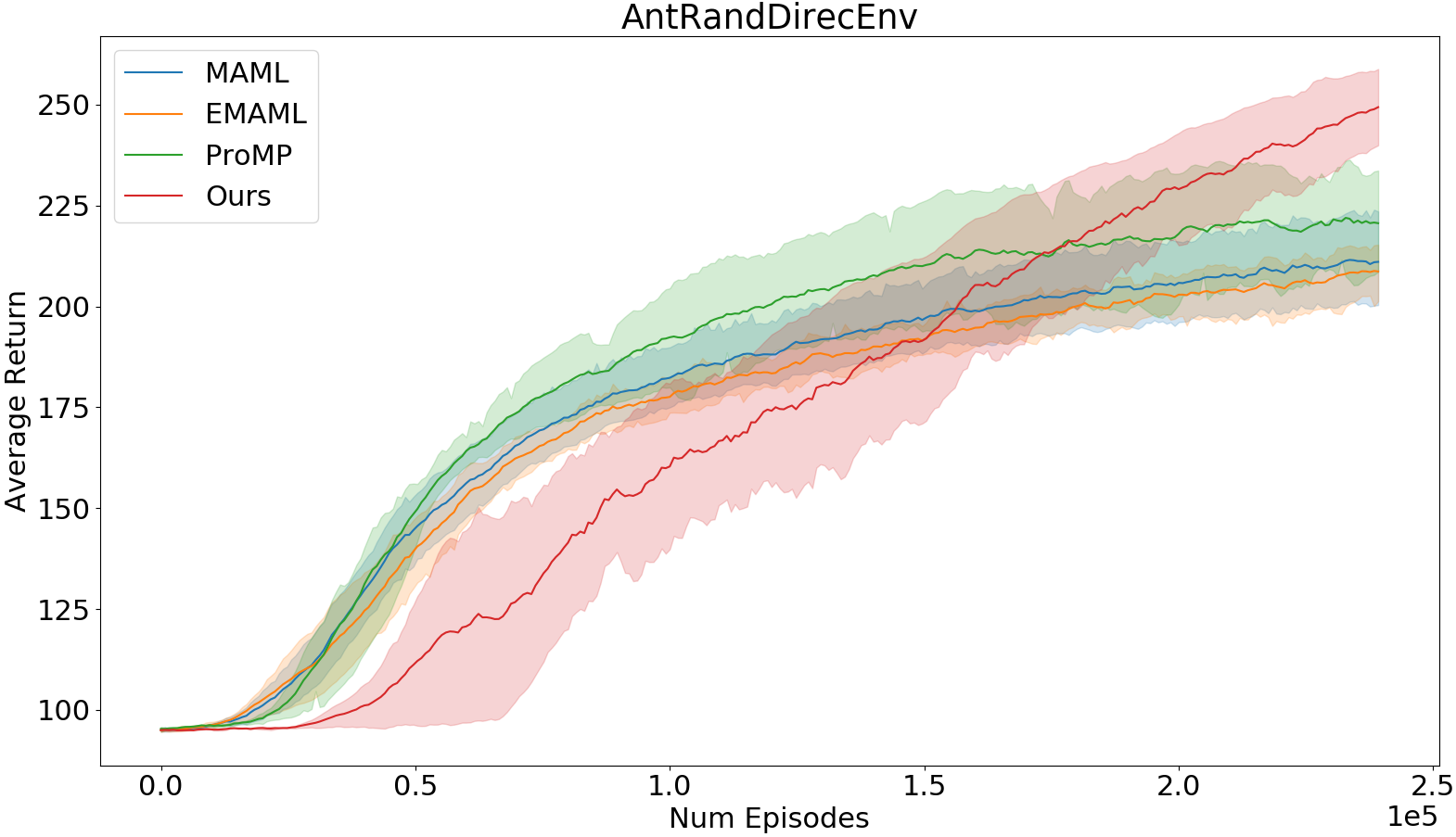} 
        \subcaption{\texttt{Ant-Fwd-Bwd}}
        \label{fig:hcvel}
    \end{minipage}
    \caption{Comparison of our method with $3$ baseline methods on the Meta-RL Benchmark tasks.}\label{fig:results}
\end{figure}
\vspace{-10pt}
\subsection{Meta RL Benchmark Continuous Control Tasks.}
\vspace{-5pt}

The continuous control tasks require adaptation either across reward functions (\texttt{Ant-Fwd-Bwd}, \texttt{Half-Cheetah-Fwd-Bwd}, \texttt{Half-Cheetah-Vel}, \texttt{Walker2D-Fwd-Bwd}) or across dynamics (\texttt{Walker2D-Rand-Params} and \texttt{Hopper-Rand-Params}). We set the horizon length to be $100$ in \texttt{Ant-Fwd-Bwd} and \texttt{Half-Cheetah-Fwd-Bwd} environments and $200$ in others in accordance with the practice in \cite{rothfuss2018promp}. The performance plots for all the $4$ algorithms are shown in Fig. \ref{fig:results}. In all the environments, our proposed method outperforms or achieves similar performance to other method in terms of asymptotic performance.

Our algorithm performs particularly well in \texttt{Half-Cheetah-Fwd-Bwd}, \texttt{Half-Cheetah-Vel}, \texttt{Walker2D-Fwd-Bwd} and \texttt{Ant-Fwd-Bwd} environments where the $N$-step returns are informative. In \texttt{Ant-Fwd-Bwd} and \texttt{Half-Cheetah-Fwd-Bwd} environments, although we reach similar asymptotic performance as ProMP, the convergence is slower in the initial stages of training. This is because training multiple networks together can make training slower especially in the initial stages of training especially when the training signal isn't strong enough. Note that in \texttt{Walker2D-Rand-Params} and \texttt{Hopper-Rand-Params} environments, although our method converges as well as the baselines, it doesn't do much better in terms of peak performance. 
This could be attributed to the selection of the self-supervision signal. A more appropriate self-supervision signal for these environments would be the next state or successor state prediction since the task distribution in these environments corresponds to the variation in model dynamics and not just reward function. This shows that the choice of the self-supervision signal plays an important role in the model's performance. To further understand these design choices we perform some ablations on a toy environment in section \ref{sec:ablations}.

\begin{wrapfigure}{r}{0.37\textwidth}
\vspace{-10pt}
        \centering
        \includegraphics[width=0.37\textwidth]{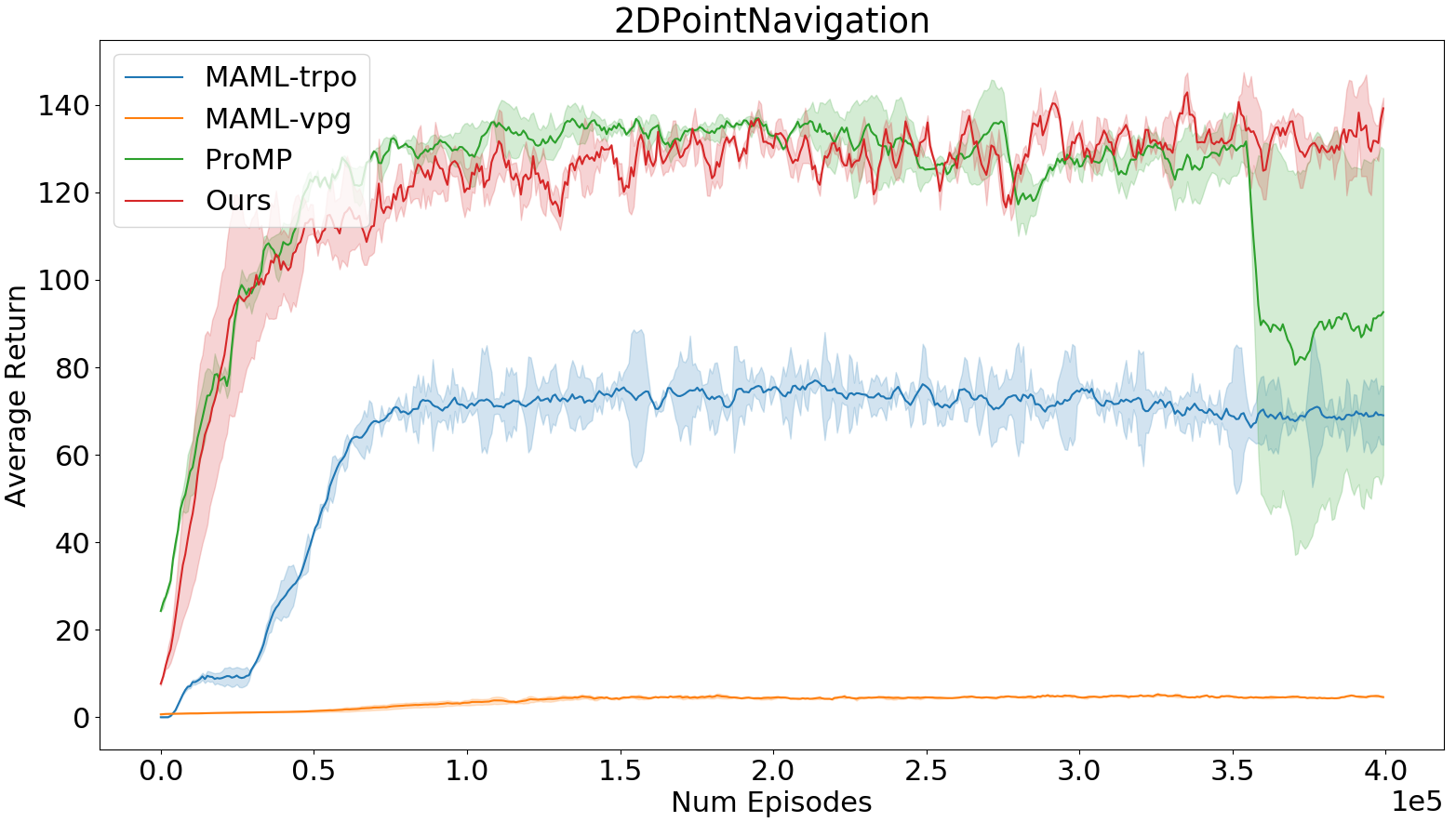} 
        \caption{2D Point Navigation}
        \label{fig:pointenv}
        \vspace{-10pt}
\end{wrapfigure}
\vspace{-10pt}
\subsection{2D Point Navigation.}
\vspace{-5pt}
We show the performance plots for ProMP, MAML-TRPO, MAML-VPG and our algorithm in the sparse reward \texttt{2DPointEnvCorner} environment (proposed in \cite{rothfuss2018promp}) in Fig. \ref{fig:results}. Each task in this environment corresponds to reaching a randomly sampled goal location (one of the four corners) in a $2$D environment. This is a sparse reward task where the agent receives a reward only if it is sufficiently close to the goal location. In this environment, the agent needs to perform efficient exploration and use the sparse reward trajectories to perform stable updates, both of which are salient aspects of our algorithm.

Our method is able to achieve this and reaches peak performance while showing stable behavior. ProMP, on the other hand, also reaches the peak performance but shows more unstable behavior than in the dense reward scenarios, although it manages to reach similar peak performance to our method. The other baselines struggle to do much in this environment since they do not explicitly incentivize exploration for the pre-update policy.

\begin{wrapfigure}{r}{0.37\textwidth}
\vspace{-25pt}
        \centering
        \includegraphics[width=.37\textwidth]{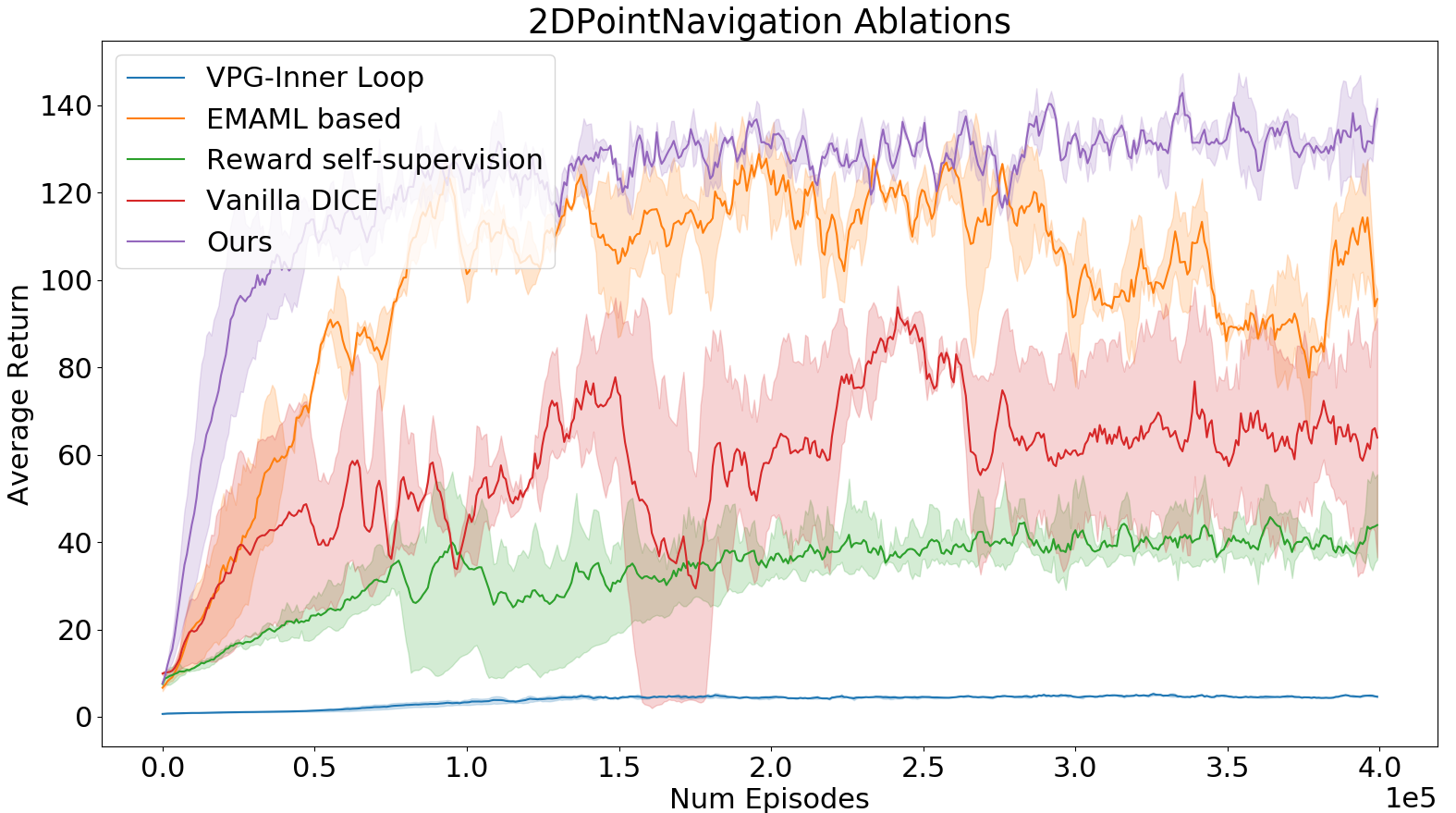}
        \caption{Ablation results}
        \label{fig:ablation}
        \vspace{-15pt}
\end{wrapfigure}
\vspace{-10pt}
\subsubsection{Ablation Study}\label{sec:ablations}
\vspace{-5pt}

We perform several ablation experiments to analyze the impact of different components of our algorithm on $2$D point navigation task. Fig. \ref{fig:ablation} shows the performance plots for the following $5$ different variants of our algorithm: 

\textbf{VPG-Inner loop}: The semi-supervised/supervised loss in the inner loop is replaced with the vanilla policy gradient loss as in MAML while using the exploration policy to sample the pre-update trajectories. This variant illustrates our claim of unstable inner loop updates when naively using an exploration policy. As expected, this model performs poorly due to the high variance off-policy updates in the inner loop. 

\textbf{Reward Self-Supervision }: A reward based self-supervised objective is used instead of return based self-supervision, i.e, the self-supervision network $M$ now predicts the reward instead of the $N$-step return at each time step. This variant is stable but struggles to reach peak performance since the task is sparse reward. This shows that the choice of self-supervision objective is also important and needs to be chosen carefully. 

\textbf{Vanilla DiCE}: In this variant, we directly use the DICE gradients to perform updates on $\phi$ instead of using the low variance gradient estimator. The leads to higher variance updates and unstable training as can be seen from the plots. This shows that the low variance gradient estimate has a major contribution to the stability during training. 

\textbf{E-MAML Based }: Here, we used an E-MAML \citep{stadie2018some} type objective to compute the gradients w.r.t. $\phi$ instead of using DICE, i.e, directly used policy gradient updates on $\mu_\phi$ but instead with returns computed on post-update trajectories. This variant ignores the causal credit assignment from output to inputs. Thus, the updates are of higher variance, leading to more unstable updates, although it manages to reach good performance.

\textbf{Ours }: The low variance estimate of the DICE gradients is used to compute updates for $\phi$ along with $N$-step return based self-supervision for inner loop updates. Our model reaches peak performance and exhibits stable training due to low variance updates.

\section{Discussions and Conclusion}
Unlike conventional meta-RL approaches, we proposed to explicitly model a separate exploration policy for the task distribution. Having two different policies gives more flexibility in training the exploration policy and also makes adaptation to any specific task easier. The above experiments illustrate that our approach provides more stable updates and better asymptotic performance as compared to ProMP when the pre-update and post-update policies are very different. Even when that is not the case, our approach matches or surpasses the baselines in terms of asymptotic performance. More importantly, this shows that in most of these tasks, separating the exploration and exploitation policies can yield better performance if properly done. From our ablation studies, we show that the self-supervised objective plays a huge role in improving stability of the updates and the choice of the self-supervised objective can be critical in some cases (e.g, predicting reward v/s return). Further, we also show through the above experiments that the variance reduction techniques used in the objective of exploration policy is important for achieving stable behavior. 
However, we would like to emphasize that the idea of using a separate exploration and exploitation policy is much more general and doesn't need to be restricted to MAML. Given the requirements of sample efficiency of the adaptation steps in the meta-learning setting, exploration is a very crucial ingredient and has been vastly under explored. Thus, we would like to explore the following extensions as future work:

\begin{itemize}
    \item Explore other techniques of self-supervision that can be more generally used across environments and tasks.
    \item Decoupling the exploration and exploitation policies allows us to perform off-policy updates. Thus, we plan to test it as a natural extension of our approach.
    \item Explore the use of having separate exploration and exploitation policies in other meta-learning approaches.
\end{itemize}

\section*{Acknowledgments} 
This work has been funded by AFOSR award FA9550-15-1-0442 and AFOSR/AFRL award FA9550-18-1-0251. We would like to thank Tristan Deleu, Maruan Al-Shedivat, Anirudh Goyal and Lisa Lee for their insightful and fruitful discussions and Tristan Deleu, Jonas Rothfuss and Dennis Lee for opensourcing the repositories and result files.
\clearpage

\bibliography{references}  

\begin{thebibliography}{27}
\providecommand{\natexlab}[1]{#1}
\providecommand{\url}[1]{\texttt{#1}}
\expandafter\ifx\csname urlstyle\endcsname\relax
  \providecommand{\doi}[1]{doi: #1}\else
  \providecommand{\doi}{doi: \begingroup \urlstyle{rm}\Url}\fi

\bibitem[Silver et~al.(2017)Silver, Schrittwieser, Simonyan, Antonoglou, Huang,
  Guez, Hubert, Baker, Lai, Bolton, et~al.]{silver2017mastering}
D.~Silver, J.~Schrittwieser, K.~Simonyan, I.~Antonoglou, A.~Huang, A.~Guez,
  T.~Hubert, L.~Baker, M.~Lai, A.~Bolton, et~al.
\newblock Mastering the game of go without human knowledge.
\newblock \emph{Nature}, 550\penalty0 (7676):\penalty0 354, 2017.

\bibitem[Olivecrona et~al.(2017)Olivecrona, Blaschke, Engkvist, and
  Chen]{olivecrona2017molecular}
M.~Olivecrona, T.~Blaschke, O.~Engkvist, and H.~Chen.
\newblock Molecular de-novo design through deep reinforcement learning.
\newblock \emph{Journal of cheminformatics}, 9\penalty0 (1):\penalty0 48, 2017.

\bibitem[Zhang et~al.(2018)Zhang, Vinyals, Munos, and Bengio]{zhang2018study}
C.~Zhang, O.~Vinyals, R.~Munos, and S.~Bengio.
\newblock A study on overfitting in deep reinforcement learning.
\newblock \emph{arXiv preprint arXiv:1804.06893}, 2018.

\bibitem[Sharaf and Daum{\'e}~III(2019)]{sharaf2019meta}
A.~Sharaf and H.~Daum{\'e}~III.
\newblock Meta-learning for contextual bandit exploration.
\newblock \emph{arXiv preprint arXiv:1901.08159}, 2019.

\bibitem[Finn et~al.(2017)Finn, Abbeel, and Levine]{finn2017model}
C.~Finn, P.~Abbeel, and S.~Levine.
\newblock Model-agnostic meta-learning for fast adaptation of deep networks.
\newblock In \emph{Proceedings of the 34th International Conference on Machine
  Learning-Volume 70}, pages 1126--1135. JMLR. org, 2017.

\bibitem[Stadie et~al.(2018)Stadie, Yang, Houthooft, Chen, Duan, Wu, Abbeel,
  and Sutskever]{stadie2018some}
B.~C. Stadie, G.~Yang, R.~Houthooft, X.~Chen, Y.~Duan, Y.~Wu, P.~Abbeel, and
  I.~Sutskever.
\newblock Some considerations on learning to explore via meta-reinforcement
  learning.
\newblock \emph{arXiv preprint arXiv:1803.01118}, 2018.

\bibitem[Rothfuss et~al.(2018)Rothfuss, Lee, Clavera, Asfour, and
  Abbeel]{rothfuss2018promp}
J.~Rothfuss, D.~Lee, I.~Clavera, T.~Asfour, and P.~Abbeel.
\newblock Promp: Proximal meta-policy search.
\newblock \emph{arXiv preprint arXiv:1810.06784}, 2018.

\bibitem[Duan et~al.(2016)Duan, Schulman, Chen, Bartlett, Sutskever, and
  Abbeel]{duan2016rl}
Y.~Duan, J.~Schulman, X.~Chen, P.~L. Bartlett, I.~Sutskever, and P.~Abbeel.
\newblock Rl$^2$: Fast reinforcement learning via slow reinforcement learning.
\newblock \emph{arXiv preprint arXiv:1611.02779}, 2016.

\bibitem[Finn and Levine(2017)]{finn2017meta}
C.~Finn and S.~Levine.
\newblock Meta-learning and universality: Deep representations and gradient
  descent can approximate any learning algorithm.
\newblock \emph{arXiv preprint arXiv:1710.11622}, 2017.

\bibitem[Snell et~al.(2017)Snell, Swersky, and Zemel]{snell2017prototypical}
J.~Snell, K.~Swersky, and R.~Zemel.
\newblock Prototypical networks for few-shot learning.
\newblock In \emph{Advances in Neural Information Processing Systems}, pages
  4077--4087, 2017.

\bibitem[Sung et~al.(2018)Sung, Yang, Zhang, Xiang, Torr, and
  Hospedales]{sung2018learning}
F.~Sung, Y.~Yang, L.~Zhang, T.~Xiang, P.~H. Torr, and T.~M. Hospedales.
\newblock Learning to compare: Relation network for few-shot learning.
\newblock In \emph{Proceedings of the IEEE Conference on Computer Vision and
  Pattern Recognition}, pages 1199--1208, 2018.

\bibitem[Nichol et~al.(2018)Nichol, Achiam, and Schulman]{alex20181storder}
A.~Nichol, J.~Achiam, and J.~Schulman.
\newblock On first-order meta-learning algorithms.
\newblock \emph{CoRR}, abs/1803.02999, 2018.
\newblock URL \url{http://arxiv.org/abs/1803.02999}.

\bibitem[Gupta et~al.(2018)Gupta, Mendonca, Liu, Abbeel, and
  Levine]{gupta2018meta}
A.~Gupta, R.~Mendonca, Y.~Liu, P.~Abbeel, and S.~Levine.
\newblock Meta-reinforcement learning of structured exploration strategies.
\newblock In \emph{Advances in Neural Information Processing Systems}, pages
  5307--5316, 2018.

\bibitem[Rakelly et~al.(2019)Rakelly, Zhou, Quillen, Finn, and
  Levine]{rakelly2019efficient}
K.~Rakelly, A.~Zhou, D.~Quillen, C.~Finn, and S.~Levine.
\newblock Efficient off-policy meta-reinforcement learning via probabilistic
  context variables.
\newblock \emph{arXiv preprint arXiv:1903.08254}, 2019.

\bibitem[Mahmoudieh(2017)]{mahmoudieh2017self}
P.~Mahmoudieh.
\newblock Self-supervision for reinforcement learning.
\newblock 2017.

\bibitem[Florensa et~al.(2019)Florensa, Degrave, Heess, Springenberg, and
  Riedmiller]{florensa2019self}
C.~Florensa, J.~Degrave, N.~Heess, J.~T. Springenberg, and M.~Riedmiller.
\newblock Self-supervised learning of image embedding for continuous control.
\newblock \emph{arXiv preprint arXiv:1901.00943}, 2019.

\bibitem[Kahn et~al.(2018)Kahn, Villaflor, Ding, Abbeel, and
  Levine]{kahn2018self}
G.~Kahn, A.~Villaflor, B.~Ding, P.~Abbeel, and S.~Levine.
\newblock Self-supervised deep reinforcement learning with generalized
  computation graphs for robot navigation.
\newblock In \emph{2018 IEEE International Conference on Robotics and
  Automation (ICRA)}, pages 1--8. IEEE, 2018.

\bibitem[Pathak et~al.(2017)Pathak, Agrawal, Efros, and
  Darrell]{pathak2017curiosity}
D.~Pathak, P.~Agrawal, A.~A. Efros, and T.~Darrell.
\newblock Curiosity-driven exploration by self-supervised prediction.
\newblock In \emph{Proceedings of the IEEE Conference on Computer Vision and
  Pattern Recognition Workshops}, pages 16--17, 2017.

\bibitem[Wortsman et~al.(2018)Wortsman, Ehsani, Rastegari, Farhadi, and
  Mottaghi]{wortsman2018learning}
M.~Wortsman, K.~Ehsani, M.~Rastegari, A.~Farhadi, and R.~Mottaghi.
\newblock Learning to learn how to learn: Self-adaptive visual navigation using
  meta-learning.
\newblock \emph{arXiv preprint arXiv:1812.00971}, 2018.

\bibitem[Yang et~al.(2019)Yang, Caluwaerts, Iscen, Tan, and
  Finn]{yang2019norml}
Y.~Yang, K.~Caluwaerts, A.~Iscen, J.~Tan, and C.~Finn.
\newblock Norml: No-reward meta learning.
\newblock In \emph{Proceedings of the 18th International Conference on
  Autonomous Agents and MultiAgent Systems}, pages 323--331. International
  Foundation for Autonomous Agents and Multiagent Systems, 2019.

\bibitem[Foerster et~al.(2018)Foerster, Farquhar, Al-Shedivat, Rockt{\"a}schel,
  Xing, and Whiteson]{foerster2018DICE}
J.~Foerster, G.~Farquhar, M.~Al-Shedivat, T.~Rockt{\"a}schel, E.~P. Xing, and
  S.~Whiteson.
\newblock Dice: The infinitely differentiable monte-carlo estimator.
\newblock \emph{arXiv preprint arXiv:1802.05098}, 2018.

\bibitem[Zintgraf et~al.(2018)Zintgraf, Shiarlis, Kurin, Hofmann, and
  Whiteson]{zintgraf2018caml}
L.~M. Zintgraf, K.~Shiarlis, V.~Kurin, K.~Hofmann, and S.~Whiteson.
\newblock Caml: Fast context adaptation via meta-learning.
\newblock \emph{arXiv preprint arXiv:1810.03642}, 2018.

\bibitem[Schulman et~al.(2017)Schulman, Wolski, Dhariwal, Radford, and
  Klimov]{schulman2017proximal}
J.~Schulman, F.~Wolski, P.~Dhariwal, A.~Radford, and O.~Klimov.
\newblock Proximal policy optimization algorithms.
\newblock \emph{arXiv preprint arXiv:1707.06347}, 2017.

\bibitem[Mnih et~al.(2016)Mnih, Badia, Mirza, Graves, Lillicrap, Harley,
  Silver, and Kavukcuoglu]{mnih2016asynchronous}
V.~Mnih, A.~P. Badia, M.~Mirza, A.~Graves, T.~Lillicrap, T.~Harley, D.~Silver,
  and K.~Kavukcuoglu.
\newblock Asynchronous methods for deep reinforcement learning.
\newblock In \emph{International conference on machine learning}, pages
  1928--1937, 2016.

\bibitem[Duan et~al.(2016)Duan, Chen, Houthooft, Schulman, and
  Abbeel]{duan2016benchmarking}
Y.~Duan, X.~Chen, R.~Houthooft, J.~Schulman, and P.~Abbeel.
\newblock Benchmarking deep reinforcement learning for continuous control.
\newblock In \emph{International Conference on Machine Learning}, pages
  1329--1338, 2016.

\bibitem[Kulkarni et~al.(2016)Kulkarni, Saeedi, Gautam, and
  Gershman]{kulkarni2016deep}
T.~D. Kulkarni, A.~Saeedi, S.~Gautam, and S.~J. Gershman.
\newblock Deep successor reinforcement learning.
\newblock \emph{arXiv preprint arXiv:1606.02396}, 2016.

\bibitem[Kingma and Ba(2014)]{kingma2014adam}
D.~P. Kingma and J.~Ba.
\newblock Adam: A method for stochastic optimization.
\newblock \emph{arXiv preprint arXiv:1412.6980}, 2014.

\end{thebibliography}
\newpage
\section{Appendix}
\subsection{Algorithm}
\vspace{-10pt}
\begin{figure}[h]
        \centering
        \includegraphics[width=1.0\textwidth]{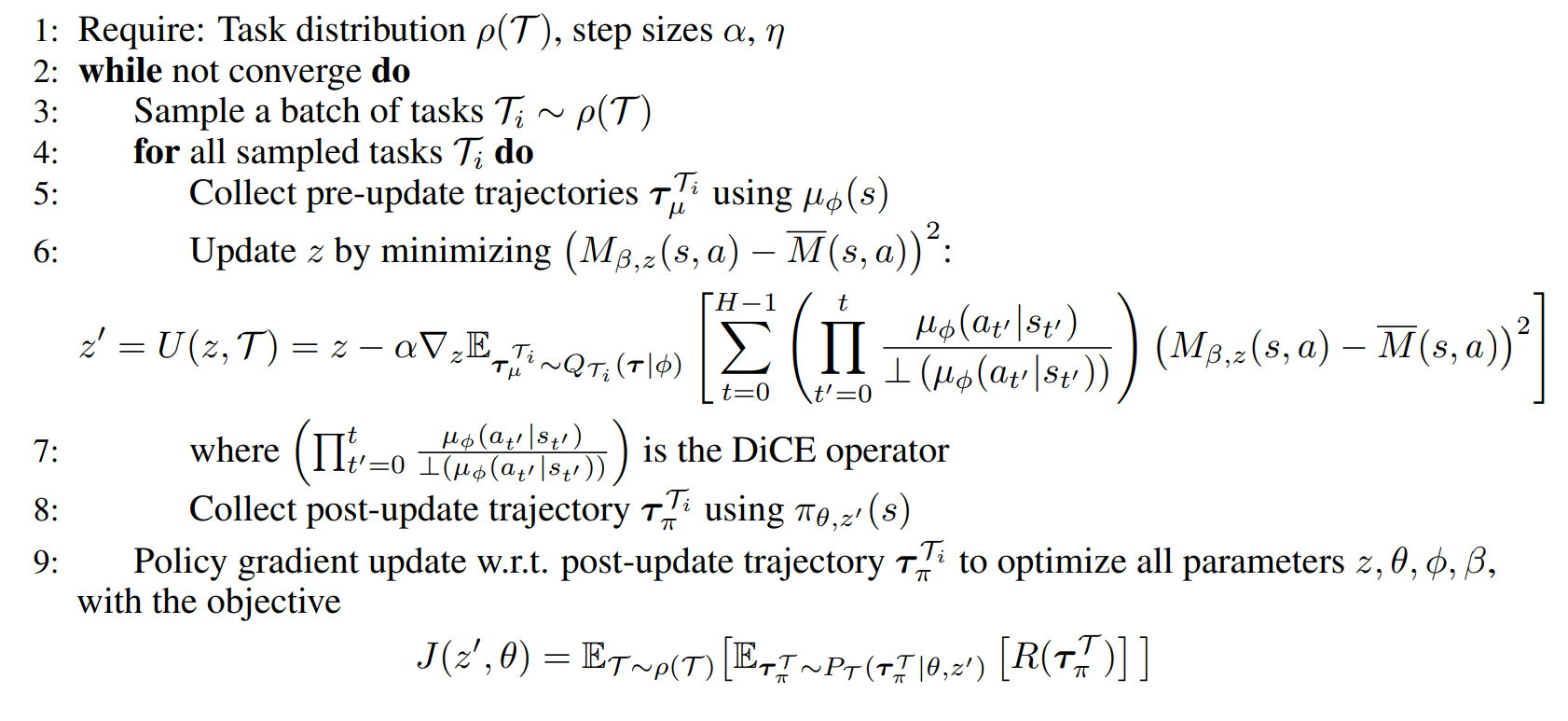}
        \label{fig:algo}
\end{figure}
\vspace{-20pt}

\subsubsection{SemiCircleEnvironment}
\begin{figure}
\begin{minipage}{.5\textwidth}
        \centering
        \includegraphics[width=\textwidth]{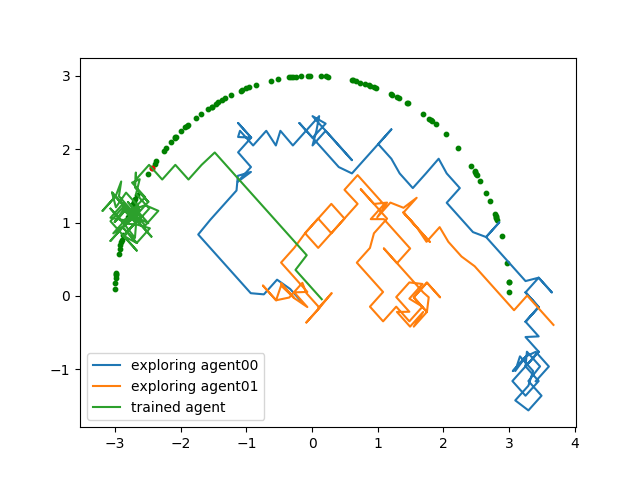}
        \caption{Ours} \label{fig:ours_semi_circle}
\end{minipage}
\begin{minipage}{.5\textwidth}
        \centering
        \includegraphics[width=\textwidth]{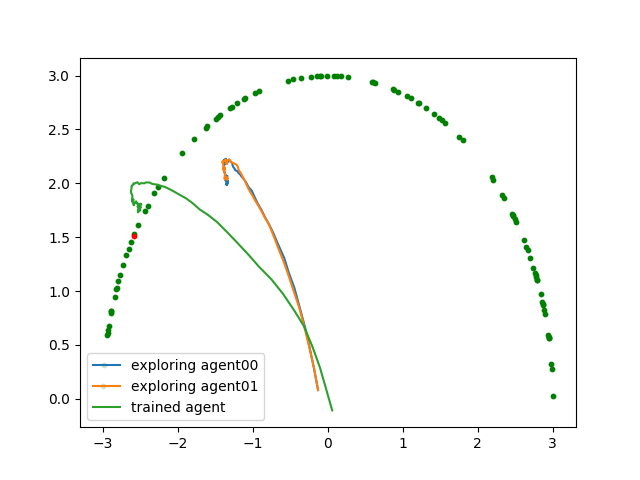}
        \caption{ProMP} \label{fig:promp}
\end{minipage}
\begin{minipage}{.5\textwidth}
        \centering
        \includegraphics[width=\textwidth]{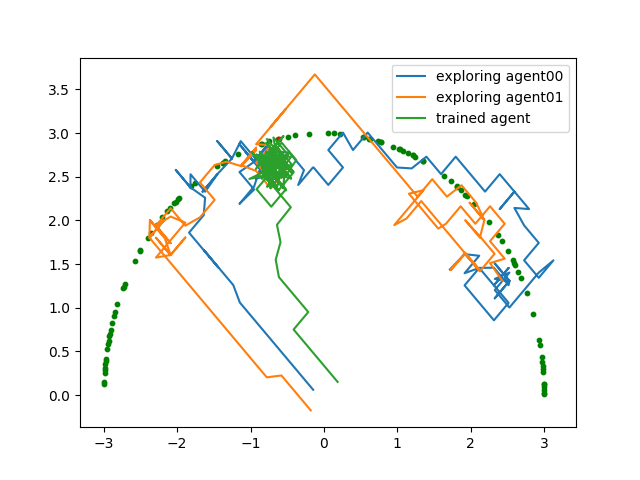}
        \caption{Ours-EnvReward : $\mu_{\phi}$ maximizes its environment rewards} \label{fig:rm_semi_circle}
\end{minipage} 
\begin{minipage}{.5\textwidth}
        \centering
        \includegraphics[width=\textwidth]{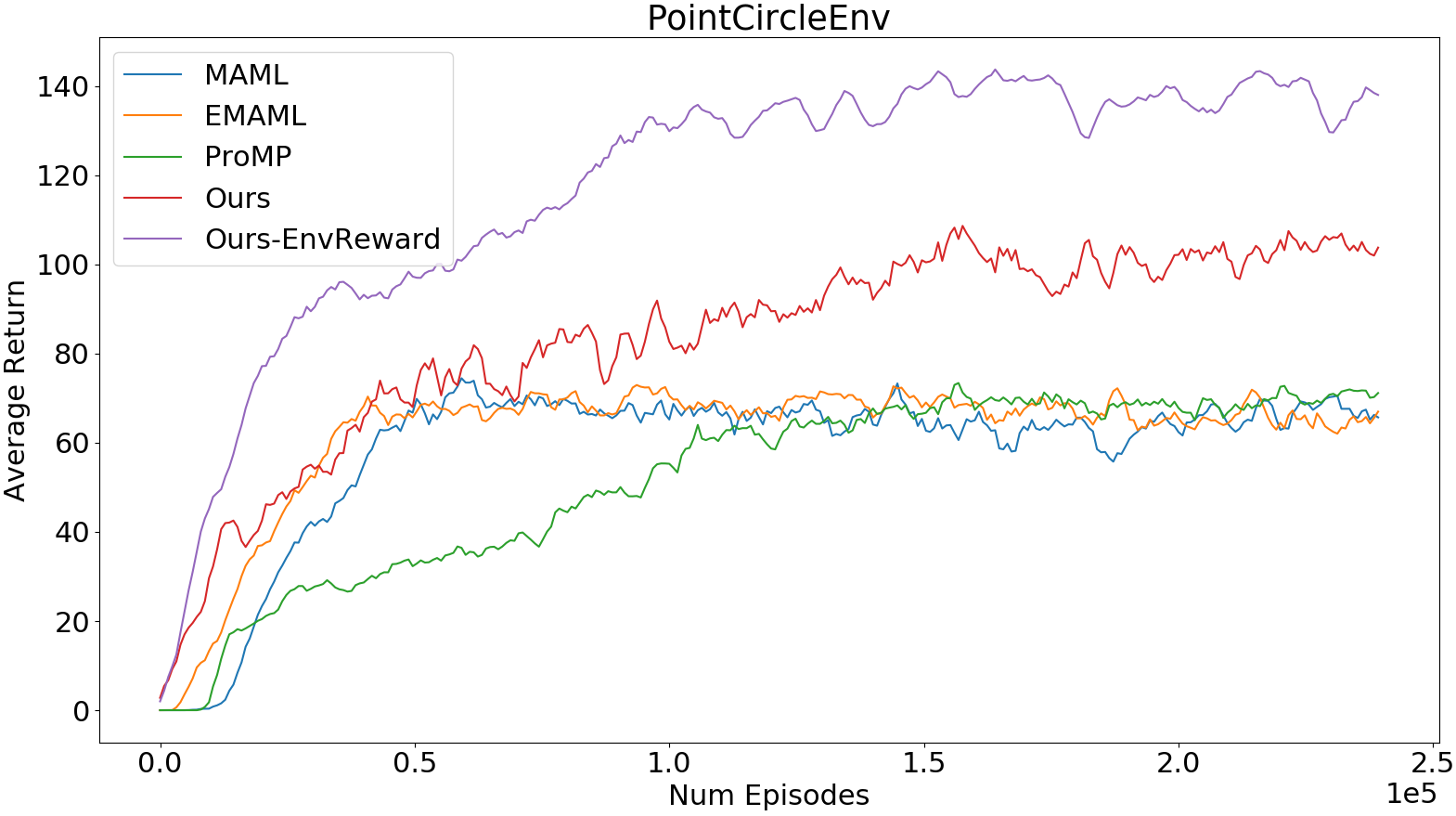}
        \caption{Comparison with baselines} \label{fig:plots_semi_circle}
\end{minipage}
\label{fig:semi-circle}
\end{figure}

We perform some additional experiments on another toy environment to illustrate the exploration behavior shown by our model and demonstrate the benefits of using different exploration and exploitation policies. Fig \ref{fig:ours_semi_circle} shows an environment where the agent is initialized at the center of the semi-circle. Each task in this environment corresponds to reaching a goal location (red dot) randomly sampled from the semi circle (green dots). This is also a sparse reward task where the agent receives a reward only if it is sufficiently close to the goal location. However, unlike the previous environments, we only allow the agent to sample 2 pre-update trajectories per task in order to identify the goal location. Thus the agent has to explore efficiently at each exploration step in order to perform reasonably at the task. Fig \ref{fig:ours_semi_circle} shows the trajectories taken by our exploration agent (orange and blue) and the exploitation/trained agent (green). Clearly, our agent has learnt to explore the environment. However, we know that a policy going around the periphery of the semi-circle would be a more useful exploration policy. In this environment we know that this exploration behavior can be reached by simply maximizing the environment rewards collected by the exploration policy. Fig. \ref{fig:rm_semi_circle} shows this experiment where the exploration policy is trained using environment reward maximization while everything else is kept unchanged. We call this variant Ours-EnvReward. We also show the trajectories traversed by promp in Fig \ref{fig:promp}. It is clear that it struggles to learn different exploration and exploitation behaviors. Fig. \ref{fig:plots_semi_circle} shows the performance of our two variants along with the baselines. This experiment shows that decoupling the exploration and exploitation policies also allows us, the designers more flexibility at training them, i.e, it allows us to add any domain knowledge we might have regarding the exploration or the exploitation policies to further improve the performance. 

\subsubsection{Varying number of adaptation trajectories collected}
\begin{figure}
\begin{minipage}{.5\textwidth}
        \centering
    \includegraphics[width=\textwidth]{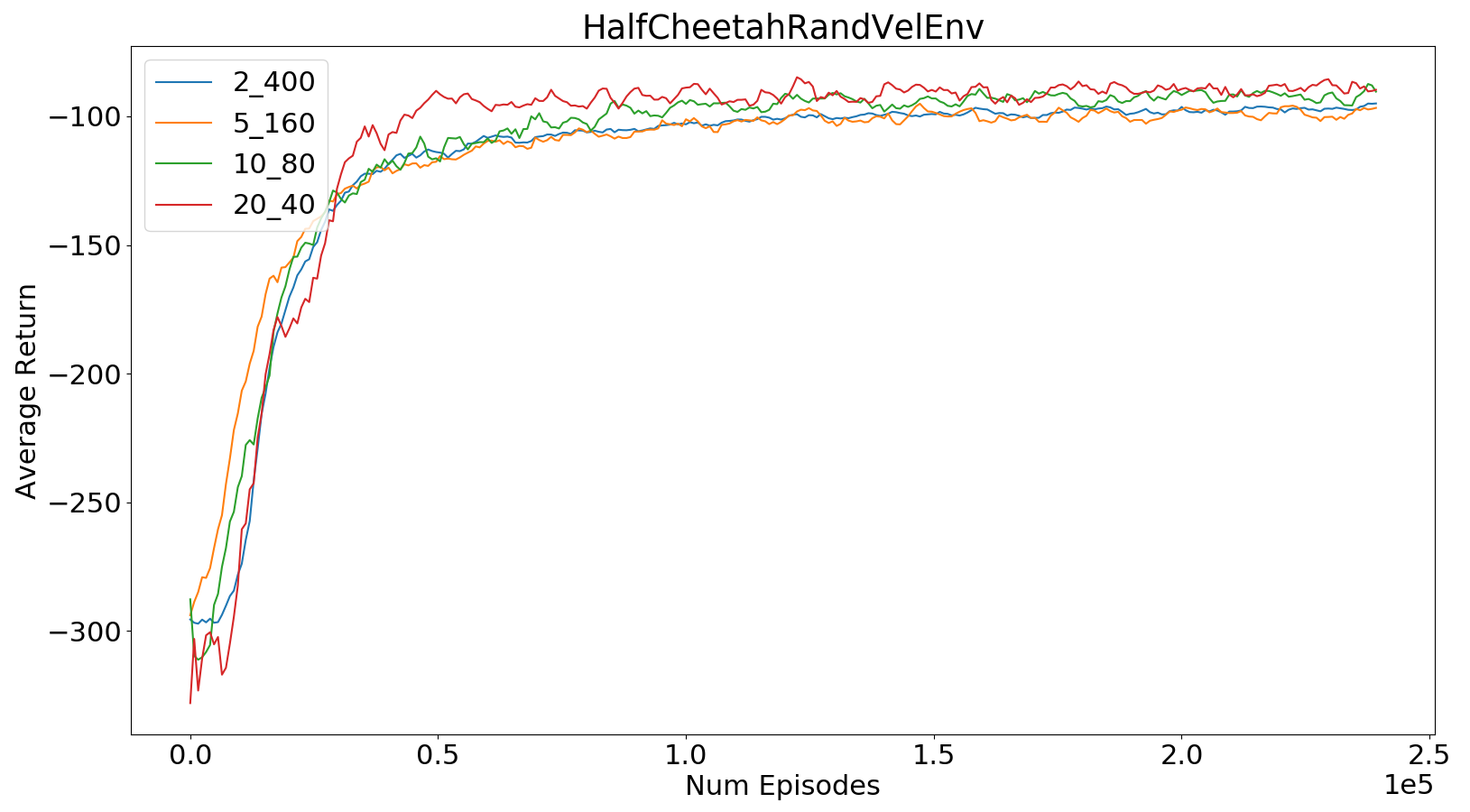}
    \caption{ProMP}
    \label{fig:my_label}
\end{minipage}
\begin{minipage}{.5\textwidth}
    \centering
    \includegraphics[width=\textwidth]{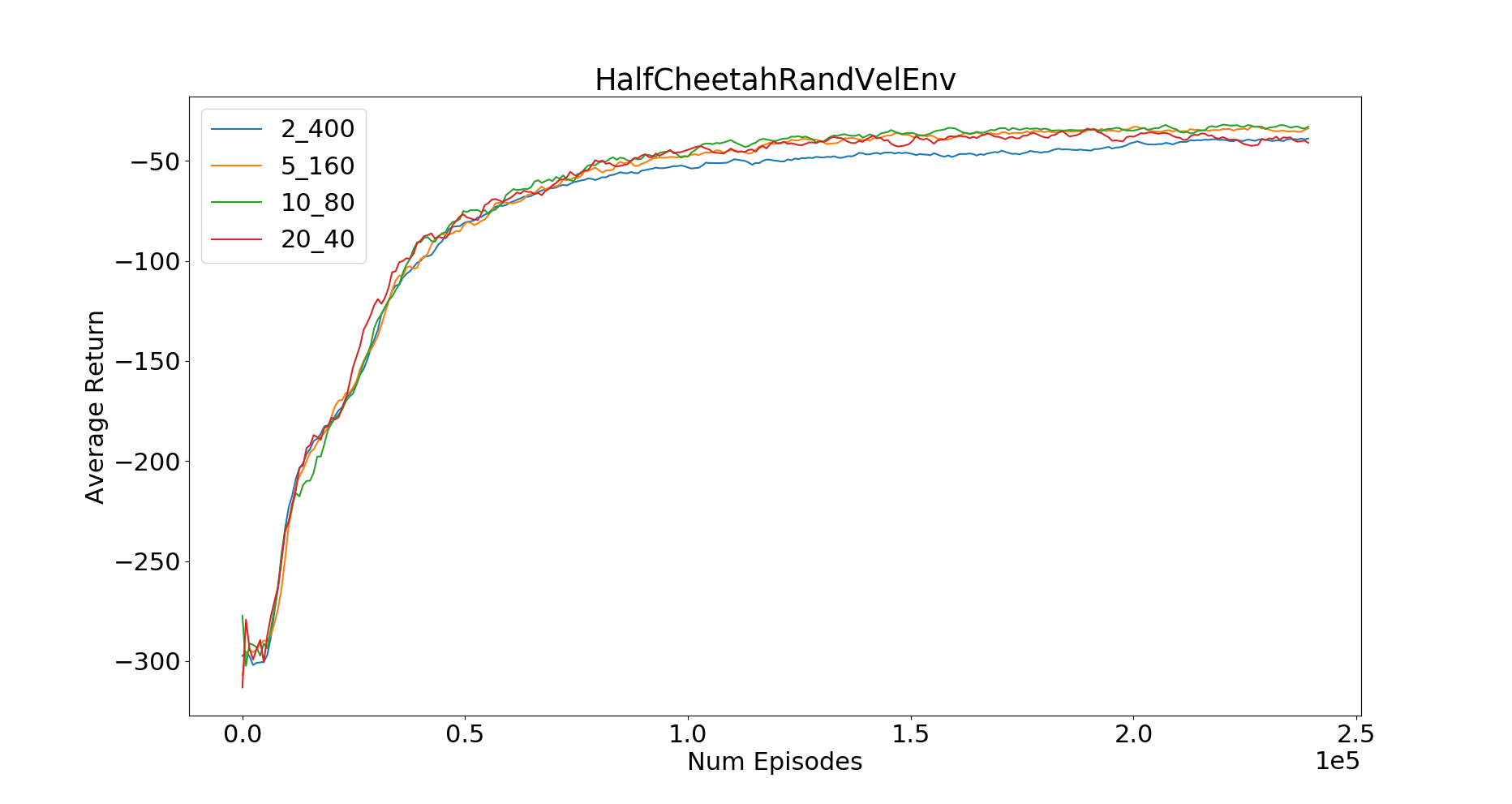}
    \caption{Ours}
    \label{fig:my_label}
\end{minipage}
\caption{Comparison with varying numbers of adaptation trajectories. In the legend "x$\_$y" corresponds to a run with x adaptation trajectories and y tasks sampled for each updated.}
\label{fig:hcv_fbs}
\end{figure}

We additionally wanted to test the sensitivity of the algorithms to the number of trajectories collected in the inner loop. This is crucial because at test time, the algorithms would only be collecting trajectories for the inner loop update, i.e, for adaptation. We test this on the HalfCheetah-Vel Environment with varying numbers of inner loop adaptation trajectories namely, 2, 5, 10 and 20. However to keep the updates stable, we increase the meta batch size (number of tasks sampled for each update) proportionally to 400, 160, 80 and 40 respectively. Figure \ref{fig:hcv_fbs} shows the plots for these variants for ProMP and our model. We notice that the performance of our model stays roughly constant across varying values of the number of adaptation trajectories whereas ProMP shows degradation in performance as the number of adaptation trajectories decrease. This shows that each of the trajectories we sample performs efficient exploration. Note that the last pair with (20,40) correspond to the standard settings of hyper-parameters which we (and other papers before us) have used for the above experiments. 

\subsection{Hyper-parameters and Details}
For all the experiments, we treat the shared parameter $z$ as a latent embedding with a fixed initial value of $\Bar{0}$. The exploitation policy $\pi_{\theta,z}(s)$ and the self-supervision network $M_{\beta, z}(s,a)$ concatenates $z$ with their respective inputs. All the three networks ($\pi, \mu, M$) have the same architecture (except inputs and output sizes) as that of the policy network in \cite{rothfuss2018promp} for all experiments. We also stick to the same values of hyper-parameters such as inner loop learning rate, gamma, tau and number of outer loop updates. We keep a constant embedding size of 32 and a constant N=15 (for computing the N-step returns) across all experiments and runs. We use the Adam~\citep{kingma2014adam} optimizer with a learning rate of $7e-4$ for all parameters. Also, we restrict ourselves to a single adaptation step in all environments, but it can be easily extended to multiple gradient steps as well by conditioning the exploration policy on the latent parameters $z$. 
\end{document}